\documentclass[10pt,conference]{IEEEtran}
\IEEEoverridecommandlockouts 
\usepackage{blkarray}                                      
\usepackage{algpseudocode}                                 
\usepackage{algorithm}
\usepackage{graphicx}                                      
\usepackage{amsmath}
\usepackage{amssymb}
\usepackage{amsfonts}
\usepackage{amsthm}
\usepackage[mathcal]{eucal}
\usepackage{mathrsfs}
\usepackage{booktabs}
\usepackage{enumerate}
\usepackage{multirow}
\usepackage[subrefformat=parens,farskip=0pt,justification=centering]{subfig}
\usepackage{color}
\usepackage{cite}                                          
\usepackage{comment}                                       
\usepackage{soul}                                          
\soulregister\cite7
\soulregister\ref7
\soulregister\pageref7
\usepackage{etoolbox}                                      
\usepackage{url}
\usepackage{nth}                                           
\usepackage{bm}                                            
\usepackage{courier}
\usepackage{balance}
\usepackage{threeparttable}
\usepackage{xcolor,colortbl}
\usepackage{footnote}
\usepackage{listings}
\usepackage{setspace}                                      
\usepackage[inline]{enumitem}

\usepackage{verbatim}
\usepackage[bookmarks=false]{hyperref}
\hypersetup{
    colorlinks = true,
    citecolor  = blue,
    linkcolor  = blue,
    urlcolor   = blue,
}
\usepackage{tikz}
\usetikzlibrary{patterns,snakes}
\usetikzlibrary{positioning,calc,fit,decorations.pathmorphing,shapes.geometric, shapes.gates.logic.US, calc}
\usetikzlibrary{arrows,arrows.meta,decorations.markings,shapes,shapes.arrows}
\usetikzlibrary{decorations,decorations.pathreplacing}
\usetikzlibrary{backgrounds}
\usepackage{filecontents}                                  
\usepackage{pgfplots}
\usepackage{pgfplotstable}
\usepackage{scalefnt}
\pgfplotsset{compat=newest}
\usepackage{caption}
\usepackage{pifont}                                        
\usepackage{cleveref}
\Crefformat{figure}{Fig.~#2#1#3}                           
\Crefname{subfigure}{Fig.}{Figs.}
\Crefname{figure}{Fig.}{Figs.}
\Crefformat{table}{TABLE~#2#1#3}                           
\usepackage[figuresright]{rotating}

\definecolor{CUHKorange}{RGB}{244,106,18} 
\definecolor{CUHKblue}{RGB}{0,111,190}    
\definecolor{CUHKgreen}{RGB}{0,127,128}   
\definecolor{CUHKred}{RGB}{228,46,36}     
\definecolor{CUHKyellow}{RGB}{198,148,34} 
\definecolor{CUHKdark}{RGB}{114,44,114}   
\definecolor{CUHKmiddle}{RGB}{144,44,144} 
\definecolor{CUHKlight}{RGB}{167,44,167} 
\definecolor{CUHKpurple}{RGB}{117,15,109}
\definecolor{CUHKgold}{RGB}{221,163,0}
\definecolor{CUHKribbon}{RGB}{244,223,176}
\definecolor{CUHKblack}{RGB}{34,24,21}

\renewcommand{\bf}[1]{\textbf{#1}}
\renewcommand{\tt}[1]{\texttt{#1}}
\renewcommand{\vec}[1]{\boldsymbol{#1}}    

\newcommand{\minisection}[1]{\vspace{.06in}\noindent{\textbf{#1}}.}

\DeclareMathOperator*{\argmax}{argmax}

\usepackage{tcolorbox}
\tcbuselibrary{skins,breakable}
    {\endtcolorbox}
    {\endtcolorbox}

\usepackage[top=0.78in,bottom=0.86in,left=0.62in,right=0.62in]{geometry}
\crefname{mytheorem}{Theorem}{Theorems}
\crefname{mylemma}{Lemma}{Lemmas}
\crefname{myclaim}{Claim}{Claims}
\crefname{myproperty}{Property}{Properties}
\crefname{mycorollary}{Corollary}{Corollaries}

\algrenewcommand\textproc{\texttt}

\makeatletter
\let\OldStatex\Statex
\renewcommand{\Statex}[1][3]{%
  \setlength\@tempdima{\algorithmicindent}%
  \OldStatex\hskip\dimexpr#1\@tempdima\relax
}
\makeatother

\RequirePackage[normalem]{ulem} 
\RequirePackage{color}\definecolor{RED}{rgb}{1,0,0}\definecolor{BLUE}{rgb}{0,0,1} 

\graphicspath{{./figs/}{../}}
\usepackage{graphicx}
\algrenewcommand\textproc{\texttt}
\usepackage{mdframed}

\begin{document}

\twocolumn

\title{
    Evolution of Optimization Algorithms for Global Placement via Large Language Models
}

\author{
    Xufeng Yao$^1$,   \
    Jiaxi Jiang$^1$,  \
    Yuxuan Zhao$^1$,  \
    Peiyu Liao$^1$,   \
    Yibo Lin$^2$,     \
    Bei Yu$^1$
    \thanks{
        This work is supported in part by The Research Grants Council of Hong Kong SAR (Project No.~CUHK14208021).
    }
    \thanks{
        The authors are with The Chinese University of Hong Kong$^1$, Hong Kong SAR. Peking University$^2$.
    }
}

\maketitle
\pagestyle{empty}

\begin{abstract}
Optimization algorithms are widely employed to tackle complex problems, but designing them manually is often labor-intensive and requires significant expertise.
Global placement is a fundamental step in electronic design automation (EDA).
While analytical approaches represent the state-of-the-art (SOTA) in global placement, their core optimization algorithms remain heavily dependent on heuristics and customized components, such as initialization strategies, preconditioning methods, and line search techniques.
This paper presents an automated framework that leverages large language models (LLM) to evolve optimization algorithms for global placement.
We first generate diverse candidate algorithms using LLM through carefully crafted prompts.
Then we introduce an LLM-based genetic flow to evolve selected candidate algorithms.
The discovered optimization algorithms exhibit substantial performance improvements across many benchmarks.
Specifically, Our design-case-specific discovered algorithms achieve average HPWL improvements of \textbf{5.05\%}, \text{5.29\%} and \textbf{8.30\%} on MMS, ISPD2005 and ISPD2019 benchmarks, and up to \textbf{17\%} improvements on individual cases.
Additionally, the discovered algorithms demonstrate good generalization ability and are complementary to existing parameter-tuning methods.

\end{abstract}

\section{Introduction}
\label{sec:intro}
Global placement is a crucial step in VLSI physical design and significantly impacts the circuit performance.
Previous solutions rely on heuristic methods such as partitioning~\cite{roy2005capo},
simulated annealing~\cite{yang2000dragon2000} and min-cut~\cite{roy2006min}  as base algorithms to minimize wirelength.
Quadratic approaches~\cite{kim2012complx,kim2012maple,lin2013polar} approximate wirelength and density using
quadratic functions.
Analytical placement is the current SOTA for VLSI placement~\cite{chan2005multilevel,chen2008ntuplace3,hsu2014ntuplace4h,lu2015eplace,cheng2018replace,zhu2018generalized}. 
Analogy to deep learning, analytical placement has two important components: optimization algorithms  and differentiable objective models. 
The task is to minimize the objective, i.e., wirelength model~\cite{naylor2001non,li2007recursive,spindler2008kraftwerk2,hsu2011tsv,hsu2013tsv,sun2019big,liao2023moreau}, via moving cell locations. The cell locations are optimized via the gradients of the objective which is largely dependent on optimization algorithms.

Due to complexity of the problem, existing optimization algorithms in placement engine remains highly heuristic and customized, including heuristic initialization, customized preconditioning methods and tailored line search techniques.
Initialization strategies determine the initial positions of macros and cells, which significantly impact the final optimization results.
Previous solutions~\cite{chen2008ntuplace3,lu2015eplace} formulate quadratic functions and solve it to determine initial placements.
Later DREAMPlace~\cite{lu2015eplace} shows that random initialization can obtain competitive results with much shorter initialization time. 
As a result, it provides a heuristic that moving all cells to the center of the layout.
Despite this, macro initialization remains challenging due to its greater influence on the final results.
SkyPlace~\cite{Jaekyung2024SkyPlace} proposes macro-aware clustering and semidefinite programming relaxation for placement initialization, which consumes longer running time overhead due to the SDP solver cost.
Preconditioning is another critical part in optimization algorithms~\cite{kim2012complx,viswanathan2007fastplace,lu2015eplace}, which  usually aims at solving the inverse matrix of the Hessian matrix for  optimization. 
For instance, ePlace~\cite{lu2015eplace} approximates the inverse of Hessian matrix using vertex and net degrees in nonlinear optimization.
Lastly, gradient-based optimizers form the foundation of analytical approaches, often with customized components. 
For instance, DREAMPlace families~\cite{lin2019dreamplace,lin2020dreamplace,gu2020dreamplace,liao2022dreamplace,chen2023stronger} introduces a Barzilai-Borwein method enabled Nesterov
algorithm, and other heuristics include noise injection at high overflow plateaus and placement termination based on divergence checks. 
Overall, designing algorithm remains challenging due to the NP-hard nature of the problem~\cite{garey1974some}.


\begin{figure}[tb!]
    \centering
    \includegraphics[width=1.\linewidth]{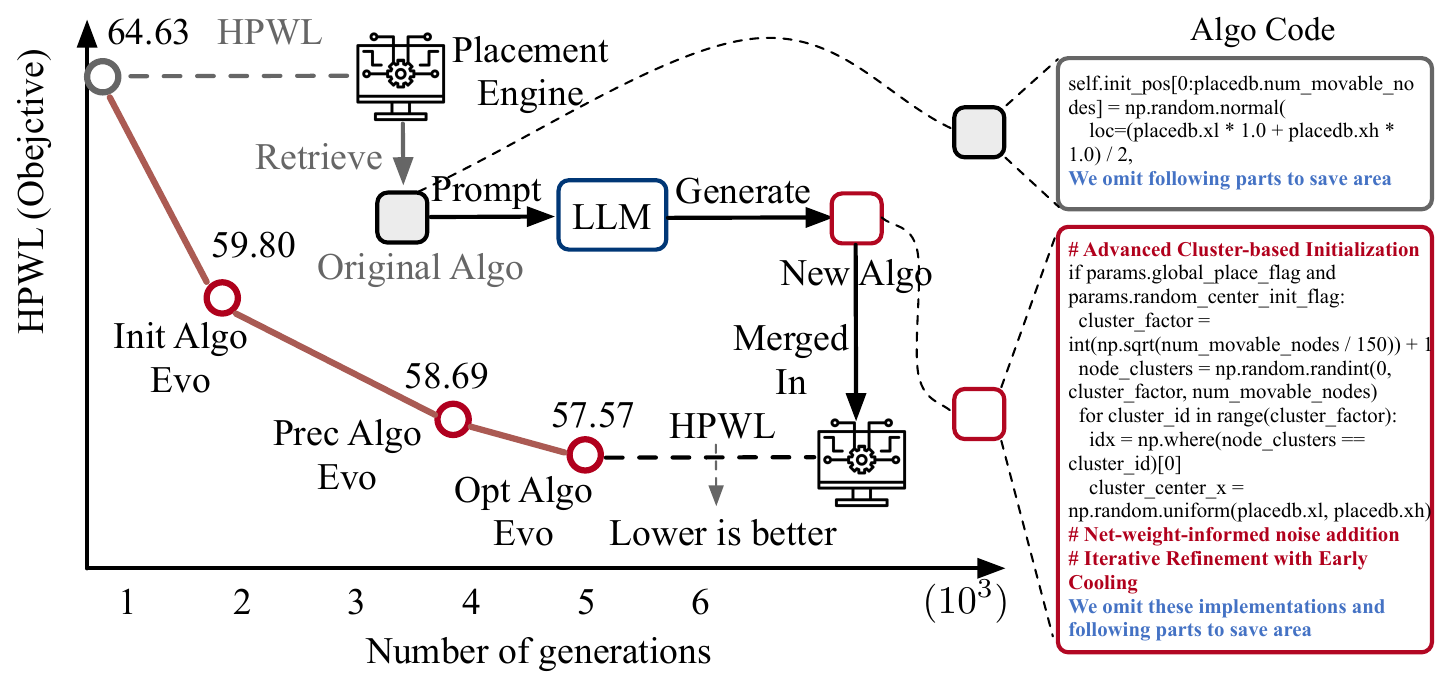}
    \caption{Evolution of Optimization Algorithms}
    \label{figs:introduction}
\end{figure}

\textbf{Key Motivation:}
Large language models (LLM) have shown great potential in many areas~\cite{achiam2023gpt,liu2023chipnemo,liu2023verilogeval,thakur2023benchmarking,fu2023gpt4aigchip,liu2024rtlcoder,saha2024llm,chang2024data,lai2024analogcoder}.
In the field of automatic algorithm design, recent works~\cite{romera2024mathematical,liu2024systematic} have shown that LLMs can evolve algorithms tailored to specific combinatorial optimization problems~\cite{seiden2002online}.
The key motivation is to leverage the advanced capacity of LLM to comprehend the problem, provide various heuristic ideas, implement these ideas into code and improve the existing algorithm code via execution feedback.
Given their remarkable performance, a key question arises:
\textbf{Can LLM improve existing optimization algorithms for global placement?} 
Nevertheless, current works mainly addresses small-scale optimization problems, leaving industry challenges, like placement, largely unexplored.
To address this challenge, this paper presents an automated framework that utilizes LLM to evolve optimization algorithms for global placement. Our framework focuses on three key optimization components: initialization, preconditioner, and optimizer. 
As shown in~\Cref{figs:introduction}, the process begins by retrieving the original algorithm codes from the placement engine. 
These codes are then combined with carefully curated prompts for the LLM, which generates and refines new algorithm codes. 
The evolved algorithm codes are integrated back into the placement engine to produce the results.
The evolution framework executes this process iteratively to generate and evolve optimization algorithms.
The details of evolution framework includes algorithm candidates generation and evolution two processes. 
Specifically, we first generate large amounts of candidate algorithms offline via LLM, then we introduces an efficient LLM-based genetic flow to evolve chosen candidates.
We leverage a large GPU cluster to accelerate whole evolution process to make it within a desired running time.
Additionally, we introduce an algorithm-level design space exploration flow tailored for resource-constrained scenario. 

\begin{figure}[tb!]
    \centering
    \includegraphics[width=.78\linewidth]{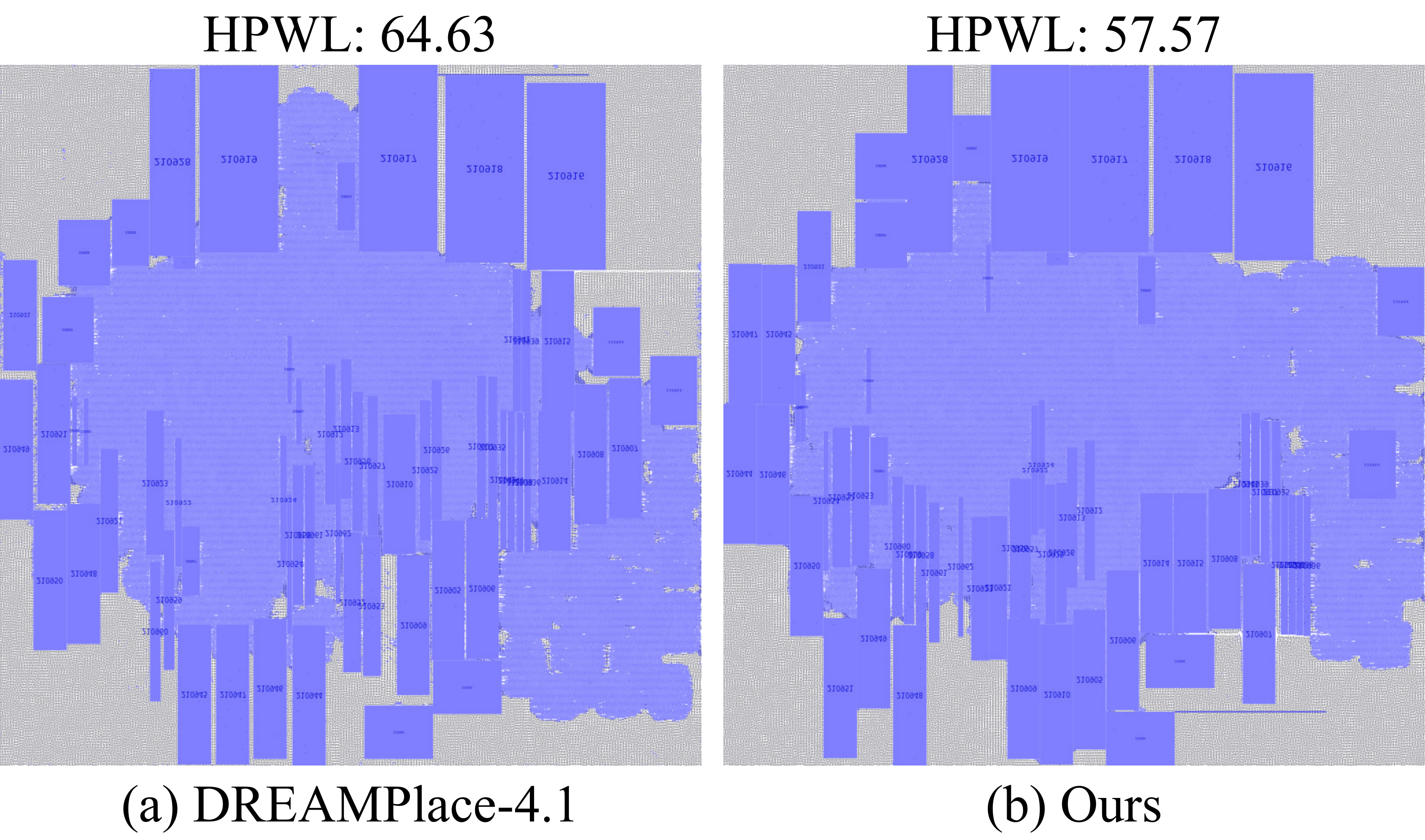}
    \caption{
        Comparison between DREAMPlace-4.1 and ours.
    }
    \label{figs:results-compare}
\end{figure}

\textbf{Key Findings: LLM exhibit remarkable capabilities in evolving optimization algorithms, resulting in significant improvements in placement quality.}
\begin{figure}[tb!]
    \centering
    \includegraphics[width=.918\linewidth]{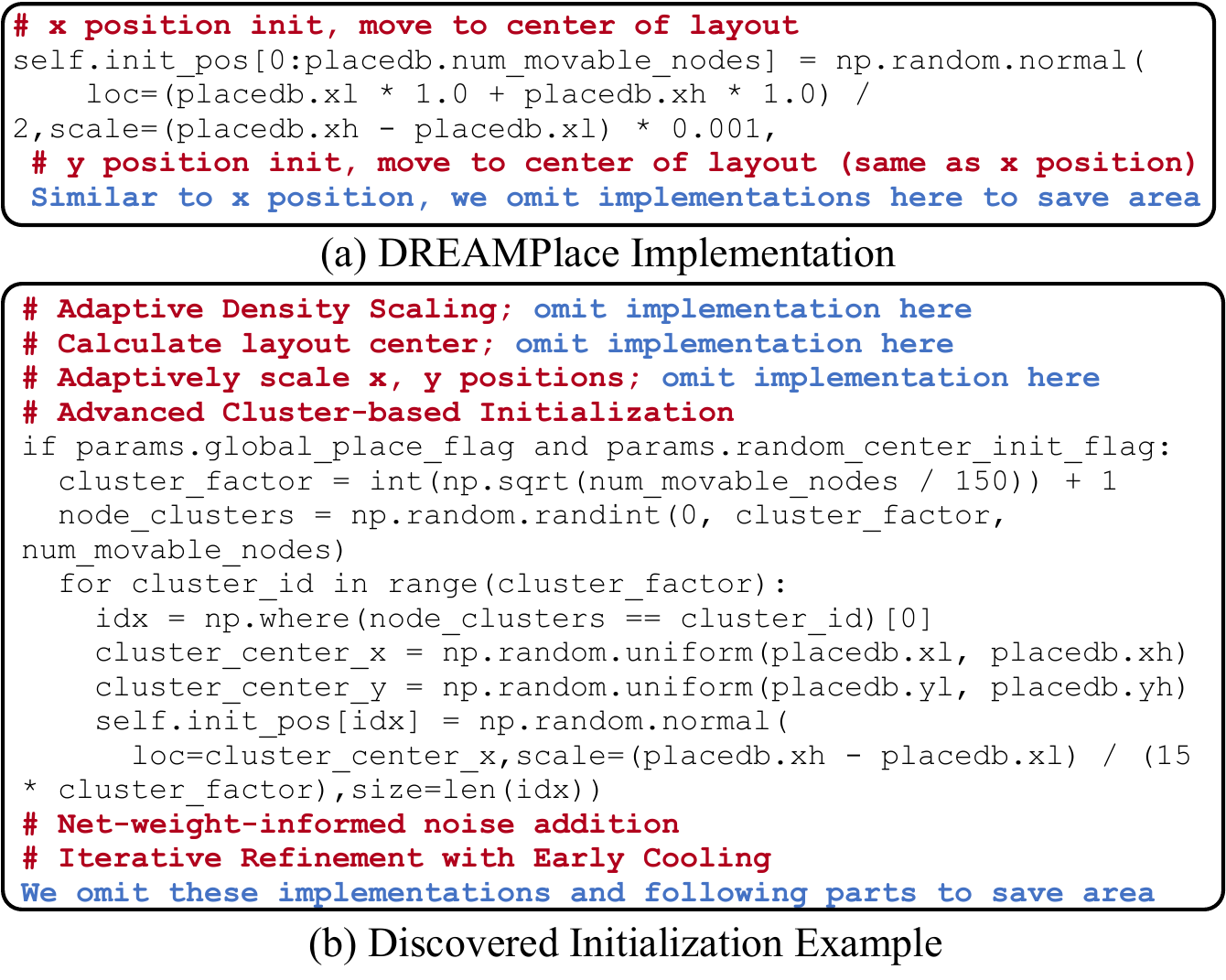}
    \caption{
        Discovered algorithms example.
    }
    \label{figs:algorithm-compare}
\end{figure}
\Cref{figs:algorithm-compare} provides an example for discovered initialization algorithm code snippet.
Compared with DREAMPlace version which simply move x, y position to the center of layout with a simple noise addition for exploration purpose.
The discovered algorithm leverages more net information and add more heuristics including adaptive density scaling, clustering, net-weight-informed noised addition and other heuristics to improve macro initialization.
\Cref{figs:results-compare} compares our method with DREAMPlace-4.1 on the MMS benchmark adaptec1 design case. Our placement appears more compact, aesthetic and regularity. Quantitatively, our approach achieves over \textbf{10\%} improvement in Half-Perimeter Wirelength (HPWL) compared to DREAMPlace-4.1. 
Note that placement is the core problem in EDA and even $1\%$ HPWL improvement is not negligible.
Our contributions are as follows:
\begin{itemize}
    \item We propose an LLM-based algorithm evolution framework to design new algorithms effectively in placement problem.
    \item Our results demonstrate significant improvement on several benchmarks under strict fair comparison, which shows the great potential of automatic algorithm design in EDA field.
    \item We release all code, prompts, and  discovered optimization algorithms, which can inspire researchers for further explorations.
\end{itemize}

\section{Related Works}
\subsection{Heuristics and Customized  Algorithms in Global Placement} \label{ssec:hs-gp}
Heuristic and customized algorithms are widely used in global placement due to the problem's NP-hard nature~\cite{garey1974some}.
For instance, AutoDMP~\cite{agnesina2023autodmp} introduces two parameters
to set the initial cell location as a percentage of the width and
height of the layout.
~\cite{sham2006optimal} introduces a macro orientation refinement heuristic.
RePlAce~\cite{cheng2018replace} introduces local and global density function with some heuristics implementations.
DREAMPlace-4.0~\cite{liao2022dreamplace} improves preconditioner by incorporating net-weighting information.
DREAMPlace-4.1~\cite{chen2023stronger}  introduces a robust SA algorithm to solve legalization issues.
RePlAce~\cite{cheng2018replace} proposes a dynamic step size adaptation method.
ePlace~\cite{lu2015eplace} introduces Nesterov optimizer with tailored line search techniques, which incorporates carefully designed optimizer parameters.
This work mainly focuses on mentioned three components while ignoring other heuristic parts. 
However, our framework can easily incorporate these heuristics and make it into a large evolution framework.

\subsection{Automatic Algorithm Design and Evolution} \label{sec:llm4algo}
Designing algorithms is labor-intensive and requires significant expertise.
Recent works that leveraging LLM to design algorithms provide new perspective for this problem~\cite{liu2024systematic}.
FunSearch~\cite{romera2024mathematical} is the pioneer paper that leveraging LLM for algorithms generation and evolution targeting optimization problem such as bin packing. 
It defines an algorithm-level code prompt template including ``evaluate'', ``solve'' and ``heuristic'' to prompt LLM to produce new heuristics and generate related code.
The generated code is executed, and feedback is used to iteratively refine the heuristics and code through the LLM.
Later several works follow this pattern and improve by introducing sophisticated prompt engineering such as evolve both heuristic and code~\cite{liu2024evolution}, exploring multi-objective setting~\cite{yao2024multi} and designing new cost functions~\cite{yao2024evolve}.
However, current LLM for algorithm design approach does not touch complex NP-hard industry challenges like placement~\cite{garey1974some}, where LLM has to leverage placement knowledge prior to enhancing existing algorithms, such as initialization.

\section{Method}
\label{Sec:algo}

\begin{figure}[tb!]
    \centering
    \includegraphics[width=.758\linewidth]{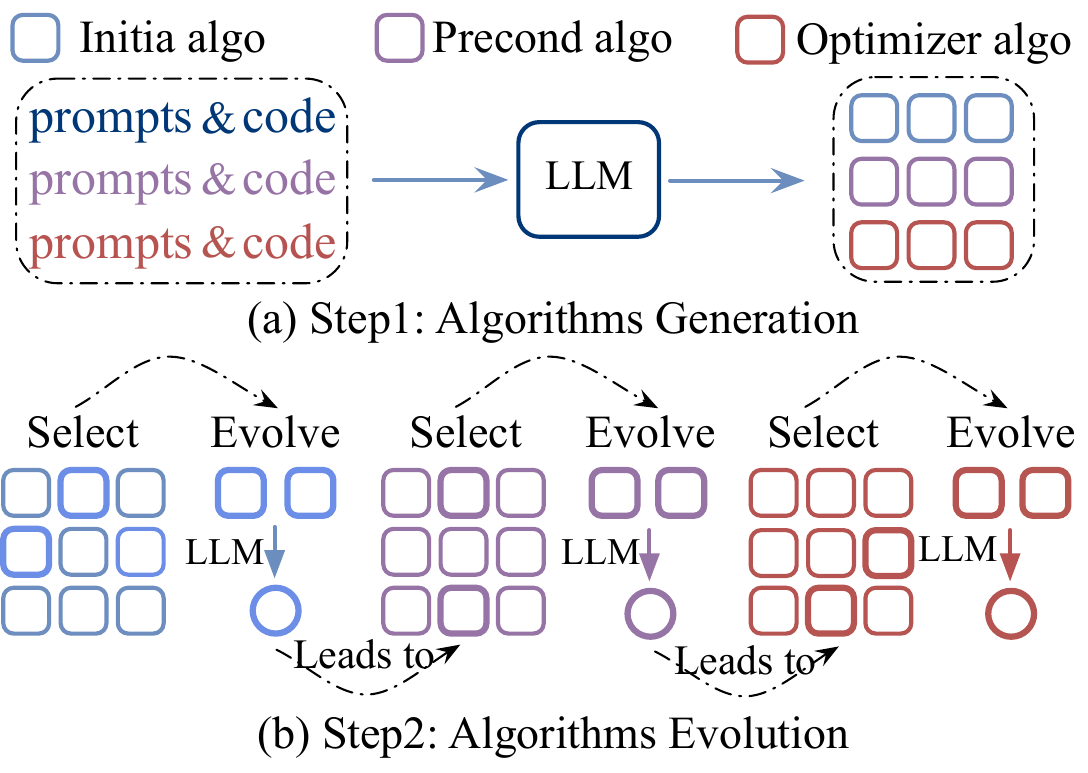}
    \caption{
        LLM-based Algorithm Generation and Evolution Pipeline.
    }
    \label{figs:framework}
\end{figure}

Our framework leverages LLM to evolve optimization algorithms for global placement task, specifically targeting three optimization components: initialization, preconditioner, and optimizer. 
The algorithm evolution process follows a sequential pipeline where each component builds upon its predecessors.
For example, the preconditioner is selected and evolved based on the evolved initialization algorithm. 
As shown in~\Cref{figs:framework}, 
our framework mainly contains two steps.
In the first step, we employ LLM to generate diverse algorithm candidates for all three componenets offline (detailed in~\Cref{ssec:candidate-gen}).
The generated candidates of each component are merged into placement engine and then evaluated in parallel on a GPU cluster.
In the second step, rather than solely selecting top-performing algorithms, we implement a balanced selection strategy considering both performance and diversity as described in~\Cref{ssec:candidate-sel}.
Finally, we apply genetic algorithms to evolve the selected candidates 
as outlined in~\Cref{ssec:candidate-evo}.


Although the proposed LLM-based algorithm design demonstrates significant improvement, 
the running time is a new bottleneck in resource-constrained scenario.
Therefore, we introduce a new algorithm-level design space (DSE) exploration to tackle the running-time problem as presented in~\Cref{ssec:algo-dse}.

\subsection{Candidates Generation} \label{ssec:candidate-gen}
The candidate algorithm generation phase leverages LLM to produce a diverse set of feasible optimization algorithms. We design specialized prompts to guide LLM in generating the desired outputs.

\begin{figure}[tb!]
    \centering
    \includegraphics[width=.818\linewidth]{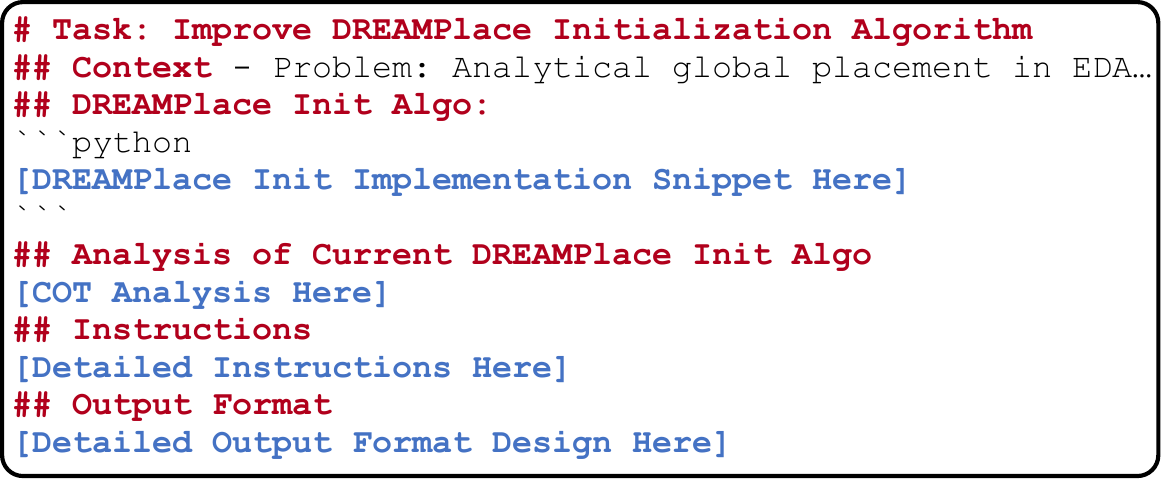}
    \caption{
        A Simplified Prompt Example.
    }
    \label{figs:prompt-example}
\end{figure}

\minisection{Prompt Construction For Optimization Algorithms}
\Cref{figs:prompt-example} demonstrates a simplified prompt template example.
We use a markdown-style format to structure the prompt, which includes task description, context, algorithm code, related analysis, specific instructions and output format. 
While the basic template structure remains fixed, components like algorithm code and related analysis are dynamically updated. The template enforces clear instructions and output formats to ensure useful and structured responses from the LLM.

\minisection{Chain-of-Thoughts Prompt Engineering}
We leverage Chain-of-Thoughts (CoT)~\cite{wei2022chain} to enhance the algorithm generation process. The LLM first analyzes the current optimization algorithm, and this analysis is then incorporated into the new prompt template. This CoT approach provides richer context for algorithm generation.
Additionally, the inherent variability in LLM's analyses introduces beneficial randomness, leading to diverse candidate algorithms.

\begin{figure}[tb!]
    \centering
    \includegraphics[width=.818\linewidth]{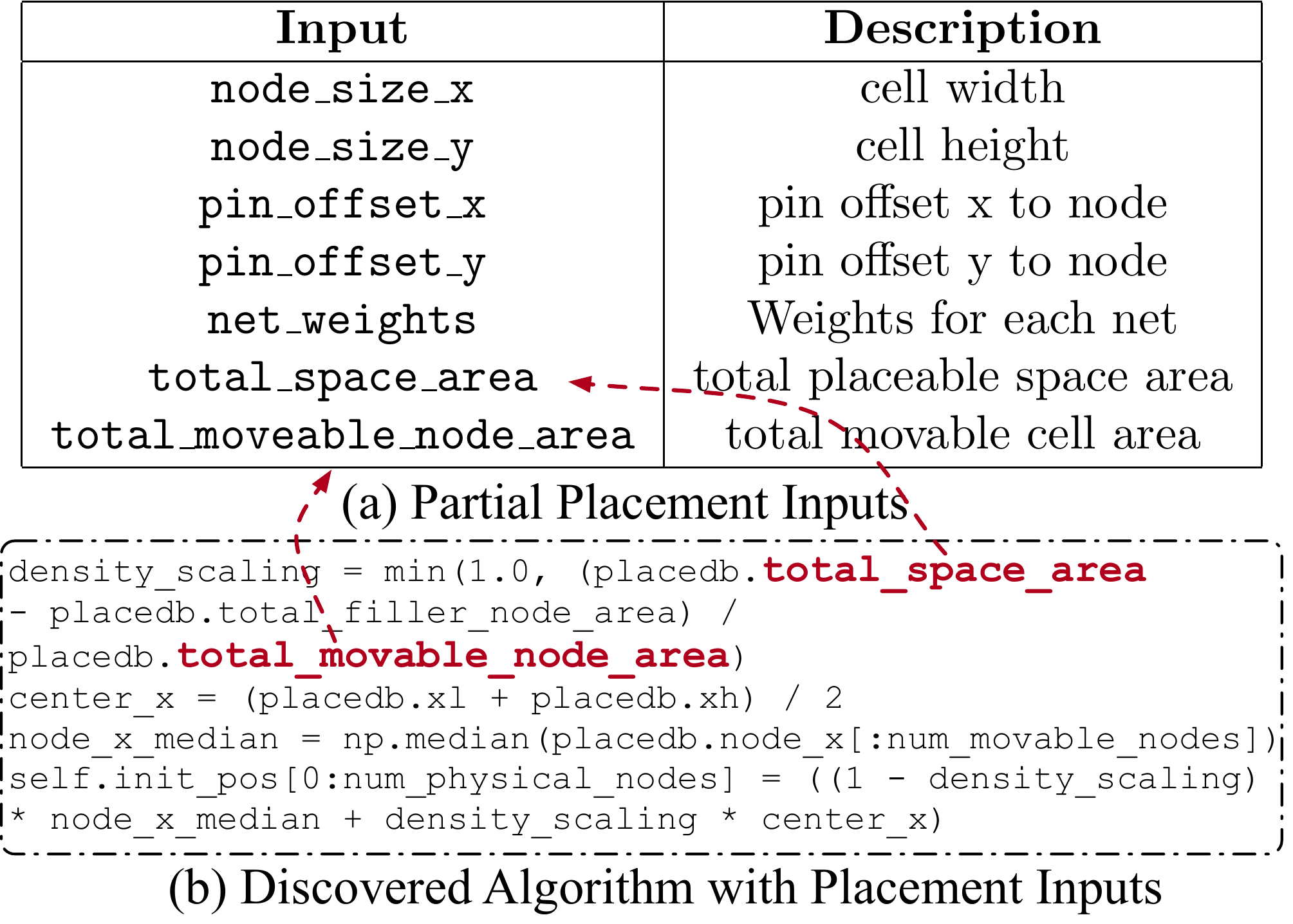}
    \caption{
        Partial Placement Inputs.
    }
    \label{figs:placement-inputs}
\end{figure}

\minisection{Leveraging Extra Placement Inputs}
Placement inputs such as logical netlist and physical cell library  contain valuable patterns that can enhance optimization algorithms, yet effective extraction and utilization of such information remains a challenge.
Despite efforts to utilize placement inputs, such as DREAMPlace-4.0~\cite{liao2022dreamplace}improving preconditioner via incorporating different net-weighting, these expert-designed algorithms still cannot fully exploit latent patterns.
We propose using LLM to dynamically select and leverage placement inputs.
~\Cref{figs:placement-inputs} shows partial placement inputs and a discovered algorithm with placement inputs example.
The discovered algorithm generated by LLM makes use of some extra placement inputs (not used in default implementation) to improve algorithms.
We collect placement inputs from the placement engine and integrate it into LLM prompts for automated feature selection and algorithm synthesis.
Results indicate that LLMs can effectively synthesize sophisticated functions to extract valuable optimization insights from placement inputs.

\minisection{Self-Referencing Prompt Engineering}
We also propose a self-referencing technique to improve candidate algorithm generation process.
First, using previous outputs, we prompt the LLM to generate a high-level idea for a new optimization algorithm. 
Next, this high-level idea guides the LLM to produce corresponding code implementation. 
Finally, this implementation serves as a reference for generating the final placement-engine-format candidate algorithms.

In general, the candidates are generated in four sequential steps: (1) LLM analyzes the given algorithm codes, (2) this analysis feeds into a new prompt template to generate high-level idea, (3) the idea guide LLM to generate reference code implementation, and (4) the reference code helps generate final desired candidate algorithms. Each step uses slightly different prompt templates with customized instructions and output formats, with details omitted for brevity.

\subsection{Candidates Selection} \label{ssec:candidate-sel}
After generating $N$ candidate algorithms, we evaluate their HPWL performance in parallel using a GPU cluster.
For the next evolution stage, we need to select top$-m$ high-performed algorithms.
The greedy approach is to select the top$-m$ algorithm with the best HPWL scores.
However, we notice that some high-performed algorithms often share similar high-level ideas, which may cause redundant LLM generation and limit exploration on other candidates' evolution.
Therefore, we propose to select top$-m$ high-performed and diverse candidate algorithms $\{\vec{a_1},\vec{a_2}, \cdots,\vec{a_m}\} \in \mathbf{A}$ by maximizing following objectives:
\begin{equation} \label{eq:1}
    \max \sum_{\vec{a_i} \in \mathbf{A}} f(\vec{a_i}) + \alpha
    \sum_{\vec{a_i},\vec{a_j} \in \mathbf{A}}
    dis(\vec{a_i}, \vec{a_j}),
\end{equation}
where $f(\cdot)$ represents the objective function that returns negative normalized HPWL value, $dis(\cdot)$ function denotes the distance function which measure the negative cosine similarity between two algorithm embeddings.
~\Cref{eq:1} is an NP-hard problem which can be reduced to the classical k-clique problem~\cite{tsourakakis2015k}. 
We provide a greedy solution to tackle this problem.
We first prune the candidate selection space by selecting top$-k$ (k $>$ m) performing candidates.
We then include the best candidate algorithm $\vec{a^*}$ into candidate lists because we observe that the best one is likely to ``win'' throughout the evolution process.
Then we iteratively select algorithms $\vec{a_i}$ into candidates lists by maximizing following objectives:
\begin{equation} \label{eq:2}
    \max \, f(\vec{a_i}) + \alpha\sum_{\vec{a_i},\vec{a_j} \in \mathbf{A} \setminus \vec{a^*}}
    dis(\vec{a_i},\vec{a_j}) + \beta \, dis(\vec{a_i}, \vec{a^*}).
\end{equation}
After each selection step, we append the chosen one into candidate lists $\mathbf{A}$.
This process ends until we collect pre-defined $m$ candidates.




\subsection{Candidates Evolution} \label{ssec:candidate-evo}
After selecting candidate algorithms, we leverage LLM to evolve chosen algorithms through an iterative process.

\minisection{Evolution Prompt Engineering}
The evolution prompt extends our candidate algorithm generation approach described in~\Cref{ssec:candidate-gen}, maintaining the use of CoT, extra placement inputs and self-reference techniques.
We enhance this process by incorporating self-reflection prompt outputs into the evolution prompt to improve the given candidate algorithms.

\minisection{Self-Reflection Prompt Engineering}
For each evolved algorithm, we integrate it into the placement engine and evaluate its HPWL performance.
The self-reflection prompt mainly deals with three possible outcomes compared with the current best candidate: (1) execution failure of the evolved algorithm, (2) HPWL improvement, or (3) HPWL degradation.
We combine the HPWL feedback into new prompt template and prompt the LLM to play a self-reflection~\cite{huang2022large} to refine the previously algorithm version.

The whole evolution process iteratively executes evolution and self-reflection prompts to obtain evolved algorithms. 
We modify instructions and output format to get desired outputs.
Denote LLM as $\pi_{\theta}$, two evolution prompts as $\vec{p_{e1}}$ and $\vec{p_{e2}}$, and reflect prompt as $\vec{p_r}$. 
$\vec{p_{e1}}$ generates an initial evolution without reflection, while $\vec{p_{e2}}$ incorporates reflection analysis.
Given $m-$th algorithm, the new evolved algorithm $\vec{a_{m+1}}$ is given by:
\begin{equation} \label{eq:evolve}
    \begin{aligned}
    \hat{\vec{a_{m}}}, \vec{r_{m}} ={}& \pi_{\theta} (\vec{p_{e1}},\vec{a_m}), \\
    \hat{\vec{r_{m}}} ={}& \pi_{\theta}(\vec{p_{r}},\vec{a_m},\hat{\vec{a_{m}}},\vec{r_m}), \\
    \vec{a_{m+1}} ={}& \pi_{\theta}(\vec{p_{e2}},\vec{a_m},\hat{\vec{a_m}},\hat{\vec{r_{m}}}),
    \end{aligned}
\end{equation}
where $\hat{\vec{a_m}}$ is the extracted evolved algorithm generated from LLM via first evolution prompt. 
$\vec{r_m}$ is the HPWL feedback of $\hat{\vec{a_m}}$, $\hat{\vec{r_m}}$ is the self-reflection analysis.
The final evolved algorithm $\vec{a_{m+1}}$ is obtained from second evolution prompt with self-reflection analysis and previous algorithm versions.
In practice, multiple evolution and self-reflection cycles can be executed in parallel for the selected candidate algorithm to improve  evolution results.

\begin{algorithm}[tb!]
\footnotesize
\caption{Evolution Flow}
\label{alg:evolution-framework}
\begin{algorithmic}[1]
    \State {\textbf{Input:}
    Sorted candidates $\mathbf{A}=\{\vec{a_1},\vec{a_2},\cdots,\vec{a_m}\}$ and iterations $T$
    };
    \State {\textbf{Output:} Best $\vec{a^*}$};
    \For {$t \leftarrow 1$ to $T$} 
    \State{Update UCB scores and choose the best $\vec{a^*}$ following~\Cref{eq:ucb}};
    \State{Evolve and Self-Reflect generated candidates $\vec{a_{m+1}}$ via~\Cref{eq:evolve}};
    \State{$\mathbf{A} \leftarrow \mathbf{A}\cup \vec{a_{m+1}}$; Sort candidates via HPWL; $\mathbf{A} \leftarrow \mathbf{A} \setminus \vec{a_{1}}$ };
    \EndFor
    \State{\textbf{return} Best $\vec{a^*}$ in candidate lists $\mathbf{A}$};
    \end{algorithmic}
\end{algorithm}

\minisection{Evolution Flow}
We select one candidate algorithm to evolve each iteration given $m$ candidate algorithms. 
To select the candidate algorithm effectively, we adopt the Upper Confidence Bound (UCB) algorithm~\cite{slivkins2019introduction}, this involves choosing the best
candidate algorithm $\vec{a^*}$ based on the evolution scores of each candidate algorithm at every step.
We define $Q(\vec{a})$ as the normalized obtained HPWL scores for a selected candidate algorithm, $N(\vec{a})$ as the number of trials for this algorithm, and $t$ as the total number of trials. The algorithms selection is based on the following equations:
\begin{equation}
    \begin{aligned}
        \vec{a^*} \leftarrow& \argmax_{\vec{a} \in \mathbf{A}} \left(Q(\vec{a}) + \lambda \sqrt{\frac{\log(t)}{N(\vec{a})}}\right).
    \end{aligned}
    \label{eq:ucb}
\end{equation}
The UCB algorithm, derived using Hoeffding’s inequality~\cite{auer2002using}, offers a sublinear regret, providing an effective balance between exploring new candidates and exploitation.

\Cref{alg:evolution-framework} demonstrates our evolution pipeline. 
Given $m$ candidate algorithms, we select one candidate each time based on their UCB score as calculated by~\Cref{eq:ucb}.
Then we evolve selected algorithms via evolution and self-reflection prompt through~\Cref{eq:evolve}.
After evolution, we append the new evolved algorithms into candidate lists. 
We sort the candidate list with HPWL values and delete the worst one.
This process iteratively evolve the candidate algorithms until it reaches the pre-defined iteration steps.


\subsection{Algorithm-level Design Space Exploration} \label{ssec:algo-dse}
\begin{figure}[tb!]
    \centering
    \includegraphics[width=.818\linewidth]{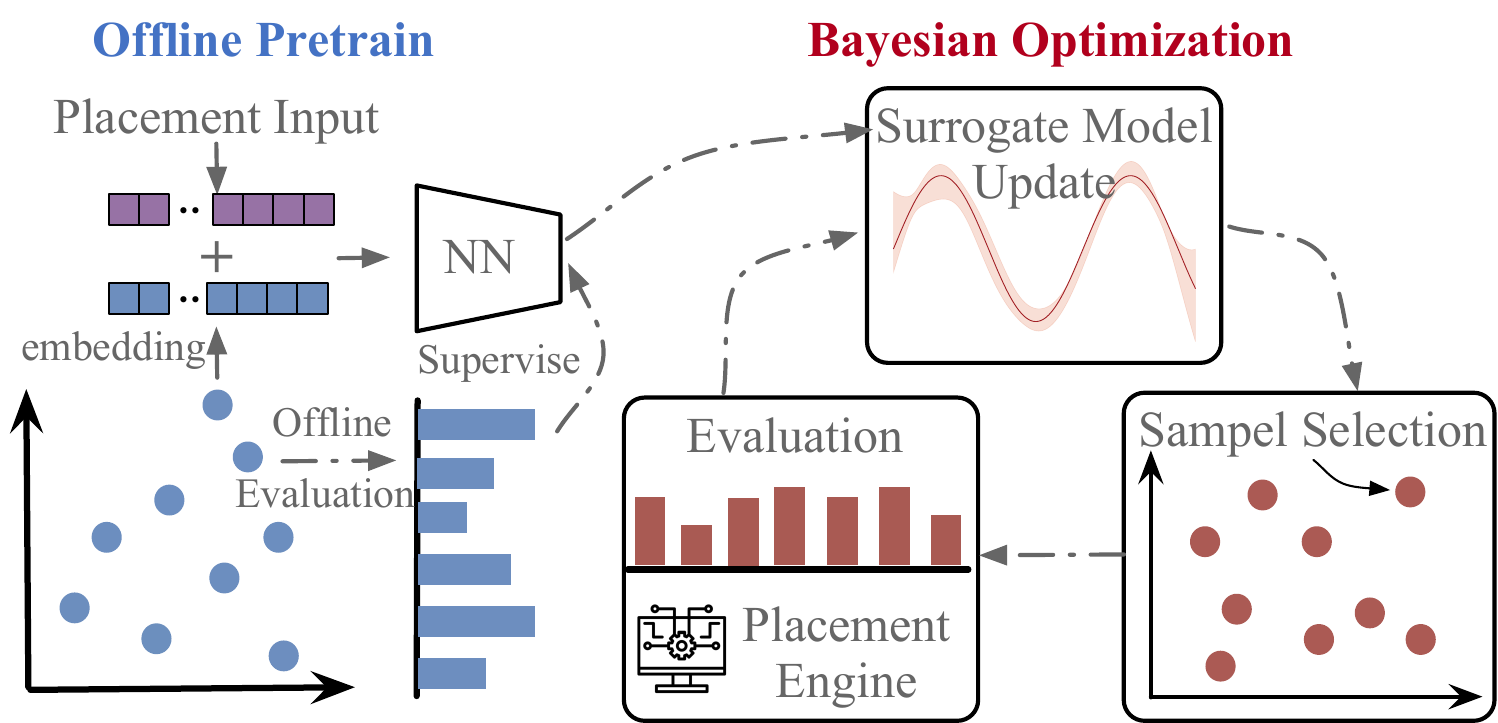}
    \caption{
        Algorithm DSE framework.
    }
    \label{figs:algo-dse}
\end{figure}

Despite the significant improvement discovered algorithms, the algorithm execution time is a potential bottleneck in resource constrained scenarios. 
Therefore, we propose an algorithm-level DSE framework to tackle this problem.
\Cref{figs:algo-dse} illustrates our Algorithm-level DSE framework, which combines offline pretraining and online search pipeline. For example, we pretrain on the ISPD2019 benchmark and leverage the surrogate model for the MMS benchmark.
We employ an embedding model~\cite{achiam2023gpt} to extract feature embeddings for candidate algorithms and a two-layer neural network to capture features from the placement inputs. These embeddings are concatenated, and a second two-layer neural network processes the combined features. The offline model is trained using HPWL results as supervision.

Similar to conventional DSE settings~\cite{zhao2017comba,schafer2019high,bai2021boom,sohrabizadeh2022autodse,agnesina2023autodmp}, we treat generated placement algorithms as design space and seek optimal solutions under time constraints (limited sampling). 
Our method employs Bayesian Optimization (BO)~\cite{snoek2012practical,lyu2018batch} with two key components: a surrogate model for performance prediction with uncertainty estimates, and an acquisition function for efficient design space exploration.
New design points are selected using the acquisition function and leveraged to update the surrogate model. 
We implement Gaussian process (GP)~\cite{williams1995gaussian} as surrogate model, characterized by mean and variance functions. 
Each time given a new data point, GP updates accordingly.
We use expected improvement (EI)~\cite{movckus1975bayesian} as acquisition function. 
\Cref{alg:dse-framework} demonstrates our Algorithm-level DSE framework.
We first randomly sample some design point from candidate lists and initialize GP and EI function.
Then we iteratively update surrogate model GP based on sampled points via EI until reaching the pre-defined iterations.
In the end, we obtain output set.

\begin{algorithm}[tb!]
\footnotesize
\caption{DSE Flow}
\label{alg:dse-framework}
\begin{algorithmic}[1]
    \State {\textbf{Input:}
    All algorithm design space $\mathbf{A}$ and optimization steps $N$
    };
    \State {\textbf{Output:} Pareto Set $\mathbf{P}$, initially $\mathbf{P} \leftarrow \emptyset$};
    \State{Randomly sample initial sets, initialize GP and EI function};
    \For {$n \leftarrow 1$ to $N$} 
    \State{Update surrogate model and EI based on sampled data};
    \State{
        $\vec{a^*} \leftarrow \arg \max_{\vec{a} \in \mathbf{A}} \text{EI}(\vec{a})$
    };
    \State{
        Run Placement Engine with $\vec{a^*}$ to obtain feedback $\vec{y}$
    };
    \State{$P \leftarrow P \cup \vec{a^*}$
    };
    \EndFor
    \State{\textbf{return}  Pareto Set from $\mathbf{P}$
    };
    \end{algorithmic}
\end{algorithm}

In this work we only consider HPWL as objective.
Therefore, we select the best evaluated algorithm in the end.
However, our framework is compatible with multi-objective optimization problem.

\section{Experimental Results} \label{sec:exp}

\begin{table*}[tb!]
    \scriptsize
    \centering
    \caption{Results on MMS benchmarks based on case-by-case algorithms}
    \label{tab:mms}
    \begin{tabular}{|c|ccc|ccc|ccc|ccccc|}
        \hline
        Design & \multicolumn{3}{c|}{Default DREAMPlace} & \multicolumn{3}{c|}{DREAMPlace w/ BB Method} & \multicolumn{3}{c|}{DREAMPlace w/ BB Method$^*$}& \multicolumn{5}{c|}{Ours} \\
        Case   & Status & HPWL & Runtime & Status & HPWL & Runtime  & Status & HPWL & Runtime  & Status & HPWL & Runtime & TT & HPWL$\downarrow$\\
        \hline \hline
        \texttt{adaptec1} & $\checkmark$               & 65.24       & 29.58  & $\checkmark$ &64.63       & 26.09  &$\checkmark$ &64.78  & 35.06  & $\checkmark$ &\bf{57.57}  & 28.75 & 0.15 & \bf{11.13\%}\\
        \texttt{adaptec2} & $\checkmark$               & 76.07       & 174.88 & $\checkmark$ &74.71       & 43.88  &$\checkmark$ &73.89  & 50.43  & $\checkmark$ &\bf{71.64}  & 49.40 & 0.38 & \bf{3.05\%} \\
        \texttt{adaptec3} & $\checkmark$               & 155.44      & 83.01  & $\checkmark$ &155.58      & 79.11  &$\checkmark$ &152.77 & 64.00  & $\checkmark$ &\bf{125.75} & 76.85 &  0.44  & \bf{17.69\%}\\
        \texttt{adaptec4} & $\checkmark$               & 142.72      & 102.79 & $\checkmark$ &142.48      & 85.94  &$\checkmark$ &141.63 & 50.93  & $\checkmark$ &\bf{119.86} & 77.82 & 0.22 & \bf{15.37\%}\\
        \texttt{adaptec5} & $\checkmark$               & 307.35      & 103.63 & $\checkmark$ &306.84      & 92.71  &$\checkmark$ &308.68 & 101.05 & $\checkmark$ &\bf{306.82} & 101.12 & 0.65 & \bf{0.59\%} \\
        \texttt{bigblue1} & $\checkmark$               & 85.16       & 39.29  & $\checkmark$ &85.32       & 32.57  &$\checkmark$ &85.25  & 32.14  & $\checkmark$ &\bf{78.64}  & 30.28 & 0.15 & \bf{7.75\%} \\
        \texttt{bigblue2} & $\checkmark$               & 125.33      & 187.13 & $\checkmark$ &125.33      & 164.33 &$\checkmark$ &124.91 & 50.33  & $\checkmark$ &\bf{124.64} & 47.80 & 0.18 & \bf{0.22\%} \\
        \texttt{bigblue3} & $\checkmark$               & 270.87      & 181.51 & $\checkmark$ &273.84      & 194.47 &$\checkmark$ &269.98 & 225.35 & $\checkmark$ &\bf{257.70} & 279.26 & 2.12 & \bf{4.55\%} \\
        \texttt{bigblue4} & $\checkmark$               & 643.55      & 233.31 & $\checkmark$ &642.88      & 282.51 &$\checkmark$ &642.02 & 247.73 & $\checkmark$ &\bf{640.74} & 246.91 & 1.46 & \bf{0.20\%} \\
        \texttt{newblue1} & $\checkmark$               & 58.61       & 32.62  & $\checkmark$ &59.33       & 34.45  &$\checkmark$ &59.52  & 38.19  & $\checkmark$ &\bf{57.47}  & 42.22 & 0.21 & \bf{3.43\%} \\
        \texttt{newblue2} & $\checkmark$               & 151.51      & 83.24  & $\checkmark$ &152.87      & 80.42  &$\checkmark$ &151.47 & 58.65  & $\checkmark$ &\bf{149.82} & 76.12 & 0.43 & \bf{1.10\%} \\
        \texttt{newblue3} & \cellcolor{red!25}$\times$ & 448.59      & 118.01 & $\checkmark$ &\bf{270.08} & 92.84  &$\checkmark$ &295.03 & 139.80 & $\checkmark$ &271.32      & 143.71 & 1.22 & \bf{8.04\%} \\
        \texttt{newblue4} & $\checkmark$               & 223.40      & 54.21  & $\checkmark$ &223.24      & 59.69  &$\checkmark$ &230.04 & 60.81  & $\checkmark$ &\bf{218.39} & 97.87 & 0.27 & \bf{5.06\%} \\
        \texttt{newblue5} & $\checkmark$               & \bf{387.90} & 132.79 & $\checkmark$ &388.98      & 160.38 &$\checkmark$ &403.11 & 191.51 & $\checkmark$ &402.97      & 204.47 & 1.62& \bf{0.03\%} \\
        \texttt{newblue6} & $\checkmark$               & 406.33      & 148.31 & $\checkmark$ &406.82      & 168.85 &$\checkmark$ &407.03 & 115.91 & $\checkmark$ &\bf{405.68} & 129.98 & 0.54 & \bf{0.33\%} \\
        \texttt{newblue7} & $\checkmark$               & 879.37      & 281.74 & $\checkmark$ &881.64      & 304.60 &$\checkmark$ &883.68 & 232.04 & $\checkmark$ &\bf{866.64} & 285.26 & 1.43& \bf{1.93\%} \\
        \hline \hline
        Ratio & & 1.000 & 1.000 & & 0.976 & 0.947 & & 0.980 & 0.919 & & \bf{0.931} & 0.995 & &\bf{5.05\%} \\
        \hline
    \end{tabular}
    
    \scriptsize{
        w/ BB Method$^*$ indicates self-implementation version based on open-source code.
        TT indicates total runtime of whole evolution process, which is measured in hours.
    }
\end{table*}

\begin{table*}[tb!]
    \scriptsize
    \centering
    \caption{Results on I/O-freed ISPD2005 benchmarks based on case-by-case algorithms}
    \label{tab:ispd2005}
    \begin{tabular}{|c|ccc|ccc|ccc|ccccc|}
        \hline
        Design & \multicolumn{3}{c|}{Default DREAMPlace} & \multicolumn{3}{c|}{DREAMPlace w/ BB Method} & \multicolumn{3}{c|}{DREAMPlace w/ BB Method$^*$}& \multicolumn{5}{c|}{Ours} \\
        Case  & Status & HPWL & Runtime & Status & HPWL & Runtime  & Status & HPWL & Runtime  & Status & HPWL & Runtime & TT & HPWL$\downarrow$ \\
        \hline \hline
        \texttt{adaptec1} & $\checkmark$               & 67.91   & 122.99 & $\checkmark$ &65.92   & 127.88 & $\checkmark$ & 72.71  & 31.34  & $\checkmark$ & 68.44       & 27.53 & 0.09 & \bf{5.87\%} \\
        \texttt{adaptec2} & $\checkmark$               & 79.74   & 145.01 & $\checkmark$ &77.65   & 143.08 & $\checkmark$ & 82.46  & 30.01  & $\checkmark$ & 80.99       & 32.55 & 0.11& \bf{1.78\%} \\
        \texttt{adaptec3} & $\checkmark$               & 153.51  & 162.94 & $\checkmark$ &151.31  & 173.44 & $\checkmark$ & 133.61 & 78.37  & $\checkmark$ & \bf{127.60} & 76.92 & 0.51 & \bf{4.50\%}\\
        \texttt{adaptec4} & $\checkmark$               & 213.86  & 86.86  & $\checkmark$ &141.45  & 85.94  & $\checkmark$ & 132.67 & 64.23  & $\checkmark$ & \bf{125.96} & 62.73 & 0.30 & \bf{5.06\%} \\
        \texttt{bigblue1} & $\checkmark$               & 83.60   & 137.89 & $\checkmark$ &82.55   & 132.94 & $\checkmark$ & 82.56  & 35.60  & $\checkmark$ & \bf{80.90}  & 35.91 & 0.15 & \bf{2.01\%} \\
        \texttt{bigblue2} & \cellcolor{red!25}$\times$ & 273.28  & 151.88 & $\checkmark$ &99.92   & 149.06 & $\checkmark$ & 98.75  & 138.47 & $\checkmark$ & \bf{91.98}  & 140.07 & 0.54 & \bf{6.86\%} \\
        \texttt{bigblue3} & $\checkmark$               & 301.70  & 161.26 & $\checkmark$ &296.79  & 197.04 & $\checkmark$ & 287.13 & 270.18 & $\checkmark$ & \bf{252.12} & 237.08 & 2.14 & \bf{12.19\%} \\
        \texttt{bigblue4} & $\checkmark$               & 658.75  & 285.61 & $\checkmark$ &620.00  & 314.12 & $\checkmark$ & 612.45 & 295.67 & $\checkmark$ & \bf{587.77} & 340.21 & 1.76 & \bf{4.03\%} \\
        \hline \hline
        Ratio & & 1.000 & 1.000 & & 0.859 & 1.043 & & 0.853 & 0.819 & & 0.809 & 0.786 & &\bf{5.29\%} \\
        \hline
    \end{tabular}
\end{table*}

\begin{table}[tb!]
    \scriptsize
    \centering
    \caption{Results on ISPD2019 based on case-by-case algorithms}
    \label{tab:ispd2019}
    {
        \begin{tabular}{|c|cc|cc|ccc|}
            \hline
            & \multicolumn{2}{c|}{ DREAMPlace}  & \multicolumn{2}{c|}{ w/ BB Method$^*$}& \multicolumn{3}{c|}{Ours} \\
            & HPWL & Runtime & HPWL & Runtime     & HPWL & Runtime  &HPWL$\downarrow$ \\
            \hline \hline
            Ratio &1.000 & 1.000 & 1.000  & 0.988 & \bf{0.906} & 1.12  &\bf{8.30\%}\\
            \hline
        \end{tabular}
    }
\end{table}

\begin{table}[tb!]
    \scriptsize
    \centering
    \caption{Results on MMS based on generalized algorithms}
    \label{tab:general-mms}
    {
        \begin{tabular}{|c|cc|cc|ccc|}
            \hline
            & \multicolumn{2}{c|}{ DREAMPlace}  & \multicolumn{2}{c|}{ w/ BB Method$^*$}& \multicolumn{3}{c|}{Ours} \\
            & HPWL & Runtime & HPWL & Runtime     & HPWL & Runtime & HPWL$\downarrow$ \\
            \hline \hline
            Ratio &1.000 & 1.000 & 0.980 & 0.919 & \textbf{0.974} & 1.01 & \textbf{0.509\%} \\
            \hline
        \end{tabular}
    }
\end{table}



\begin{table}[tb!]
    \scriptsize
    \centering
    \caption{Results on ISPD2005 based on generalized algorithms}
    \label{tab:general-ispd2005}
    {
        \begin{tabular}{|c|cc|cc|ccc|}
            \hline
            & \multicolumn{2}{c|}{ DREAMPlace}  & \multicolumn{2}{c|}{ w/ BB Method$^*$}& \multicolumn{3}{c|}{Ours} \\
            & HPWL & Runtime & HPWL & Runtime     & HPWL & Runtime & HPWL$\downarrow$  \\
            \hline \hline
            Ratio &1.000 & 1.000 & 0.853 & 0.819 & \textbf{0.824} & 0.83 & \textbf{2.85\%}\\
            \hline
        \end{tabular}
    }
\end{table} 

\begin{table}[tb!]
    \scriptsize
    \centering
    \caption{Selected results on discovered initialization algorithms}
    \label{tab:sel-results}
    {
        \begin{tabular}{|c|cc|ccc|}
            \hline
            Design   & \multicolumn{2}{c|}{ w/ BB Method$^*$}& \multicolumn{3}{c|}{Selected Init Algorithm} \\
            Case & HPWL & Runtime     & HPWL & Runtime  & HPWL$\downarrow$\\
            \hline \hline
            \texttt{adaptec1} & 64.78  & 35.06 & 61.45  & 37.95 & $5.14\%$ \\
            \texttt{adaptec3} & 152.77 & 64.00 & 129.86 & 71.71 & $14.99\%$ \\
            \texttt{adaptec4} & 141.63 & 50.93 & 123.85 & 56.48 & $12.55\%$ \\
            \texttt{bigblue1} & 85.25  & 32.14 & 78.98  & 35.65 & $7.35\%$ \\  
            \hline
        \end{tabular}
    }
\end{table}

\subsection{Implementation and Benchmarks} \label{ssec:exp-settng}
We implement our framework on top of open-sourced DREAMPlace-4.1~\cite{chen2023stronger}. To ensure reproducibility, we maintain all default hyper-parameters and settings as specified in~\cite{chen2023stronger} and fix random seeds to ensure stability of results. 



Our experiments utilize the GPT-4o API (version gpt-4o-2024-08-06) and Embedding API (version text-embedding-3-large). 
We leverage a distributed GPU cluster comprising 60 NVIDIA RTX 2080 Ti and 150 RTX 3090 GPUs for parallel execution.
We generate at least 1000 feasible candidates for each optimization components. 
In evolution phase, the total trail numbers is 1000 for each case.

We evaluate our algorithms on three benchmarks: MMS~\cite{yan2009handling}, ISPD2005~\cite{nam2005ispd2005},
and ISPD2019~\cite{dolgov20192019}. 
MMS, derived from ISPD2005/ISPD2006, is a widely-used and competitive benchmark where all macros are movable and I/O object sizes are set to zero. We also use a modified version of the ISPD2005 benchmark, where macros and I/O objects are movable without altering their shapes, as introduced in~\cite{chen2023stronger}.
ISPD2019, though originally a routing benchmark, can be adapted for placement problems, allowing us to demonstrate the generality of our framework. 
Additionally, we also observe over 5\% HPWL improvement on the TILOS benchmark~\cite{cheng2023assessment}. 
However, since some implementations in DREAMPlace-4.1~\cite{chen2023stronger} have not yet open-sourced, we refrain from releasing full results to ensure reproducibility using publicly available code.


\begin{figure}[tb!]
    \centering
    \includegraphics[width=.88\linewidth]{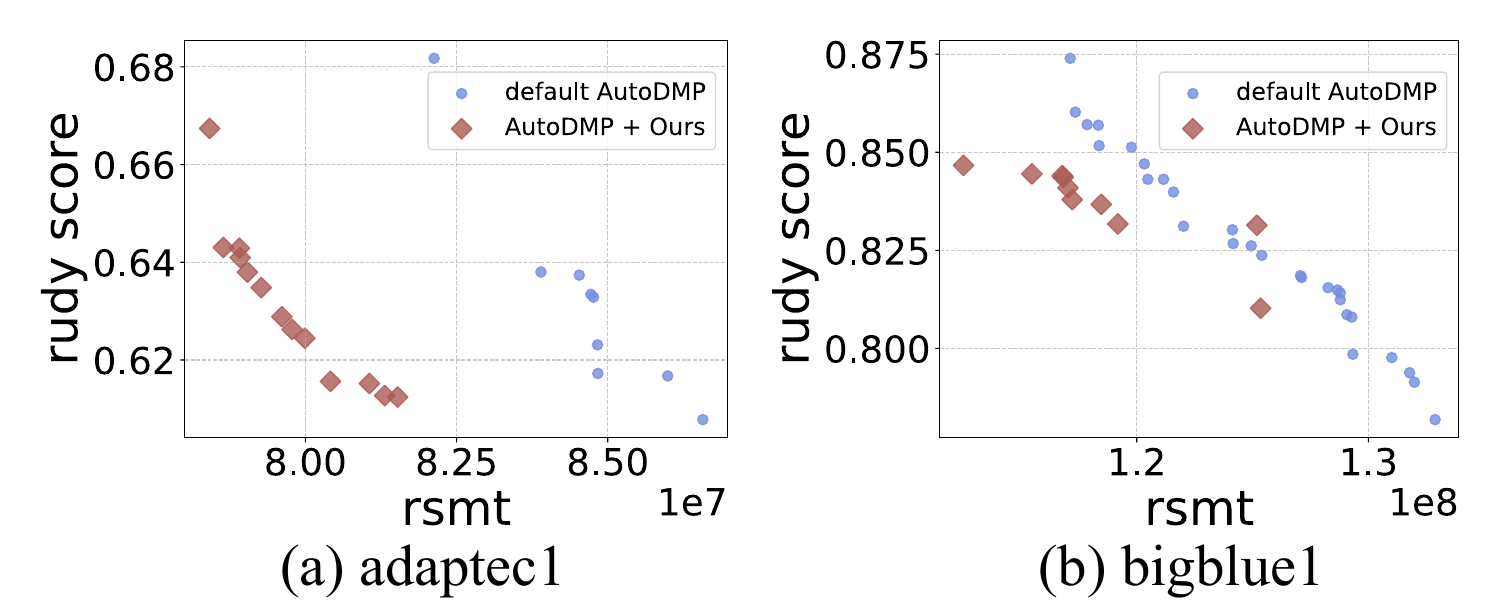}
    \caption{
       Rudy-RSMT Pareto frontiers comparison.
    }
    \label{figs:autodmp-plot}
\end{figure}

\subsection{Performance Analysis}
\minisection{Case-by-Case Results Analysis}
\Cref{tab:mms} and \Cref{tab:ispd2005,tab:ispd2019} demonstrate case-by-case evolution results.
In this setting, we evolve our algorithms via design case HPWL feedback, which is same to conventional data-driven RL methods and recent algorithm design and evolution pipeline~\cite{mirhoseini2021graph,cheng2021joint,lai2022maskplace,lai2023chipformer,shi2024macro,geng2024reinforcement,liu2024systematic,romera2024mathematical}.
Default DREAMPlace indicates DREAMPlace-4.1 without Barzilai-Borwein method which we serve as a stable baseline. 
Note that This version's performance is much better than the Default DREAMPlace performance in previous papers such as~\cite{chen2023stronger} because we use the latest open-source DREAMPlace version~\cite{chen2023stronger} with two-stage flow for mixed-size global placement and customized macro legalization. 
DREAMPlace w/ BB Method and DREAMPlace w/ BB Method$^*$ indicates the results presented on original paper~\cite{chen2023stronger} and our self-implementation results. 
We provide two version's results because case-by-case results may perform differently within different environments, sometimes even close to $1\%$ difference. 
It can be shown that our self-implement obtains similar results compared with paper result~\cite{chen2023stronger}.
Although some cases may have around $1\%$ difference, such as case newblue3 on MMS benchmark and case adaptec3 and adaptec4 on ISPD2005 benchmark. 
Nevertheless, the overall HPWL performance ratio is stable. 
We observe the discovered algorithms demonstrate significant improvement on almost all cases while maintaining same-level running-time on top of DREAMPlace engine. 
Specifically, some cases such as adaptec1, adaptec3 on MMS and biglue3 on ISPD2005 all show over $10\%$ HPWL improvement. 
Even greater gains are observed on the ISPD2019 benchmark, which, although primarily a routing benchmark, is rarely used in placement research.
We highlight the generalization capability of our algorithm-evolution framework on this benchmark.
There are two possible statuses: success ($\checkmark$) and divergence ($\times$), where divergence indicates extremely low-quality solutions.
We report the total runtime (TT) for each case, measured on GPU clusters using parallel execution. This runtime is acceptable for placement tasks and can be further reduced by adding more GPUs. \Cref{figs:autodmp-plot} presents the results of combining our discovered algorithms with the AutoDMP parameter DSE method~\cite{agnesina2023autodmp} on MMS benchmark. Our approach achieves better Rudy-RSMT Pareto frontiers than AutoDMP alone, demonstrating that our algorithms are compatible with parameter-tuning methods and can potentially improve other objectives such as RSMT.

\begin{figure}[tb!]
    \centering
    \includegraphics[width=.938\linewidth]{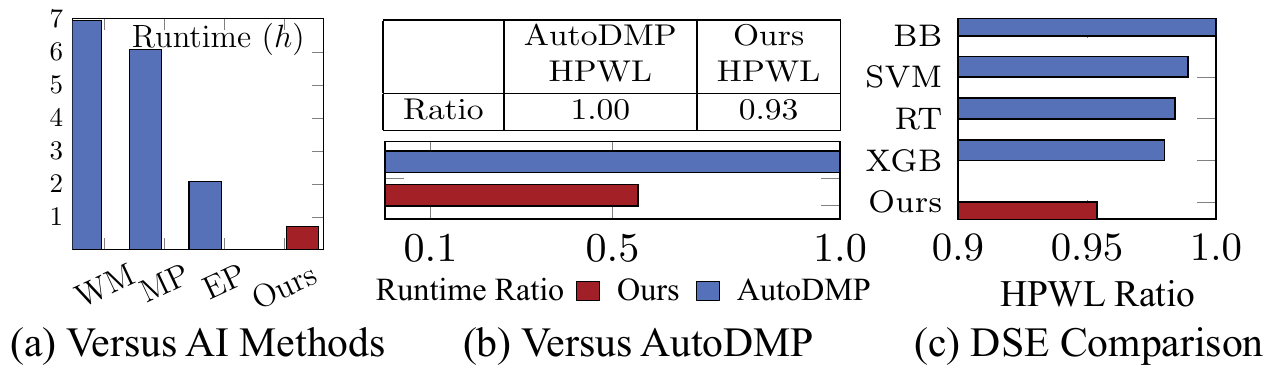}
    \caption{
        DSE comparison and running-time results.
    }
    \label{figs:dse-compare}
\end{figure}

\begin{table}[tb!]
    \scriptsize
    \centering
    \caption{DSE results on IO-freed ISPD2005 benchmarks}
    \label{tab:dse-extract}
    {
        \begin{tabular}{|c|cc|ccc|}
            \hline
              & \multicolumn{2}{c|}{ Ours (Exact)}& \multicolumn{3}{c|}{Ours (DSE)} \\
            & HPWL & Runtime     & HPWL & Runtime  & Runtime$\downarrow$\\
            \hline \hline
            Ratio &1.000 & 1.000 & 1.037 & 0.165& \textbf{83.50\%}  \\
            \hline
        \end{tabular}
    }
\end{table} 
\minisection{Generalized Results Analysis}
\Cref{tab:general-mms} and \Cref{tab:general-ispd2005} show generalized results on MMS and ISPD2005 benchmarks, where a generalized algorithm is applied to all cases.
Our generalized algorithm still achieves non-negligible improvements.
Noted that we only test on our generated candidate algorithms without evolution parts, thus the algorithm candidates do not receive any HPWL feedback.
However, although the general results are not comparable with the case-by-case evolution results, we still observe impressive generalized discovered algorithms as shown in~\Cref{figs:algorithm-compare}.
\Cref{tab:sel-results} demonstrates the selected results on MMS benchmark based on this generalized discovered algorithm.
It's clearly that the discovered algorithm achieves significant improvement across different cases. 
Note that this discovered initialization algorithm is not general for all cases and it also produce divergence performance in other cases. 
However, due to the large improvement, we believe the discovered algorithms can provide new insights for researchers.

\minisection{Runtime Analysis and DSE results}
We compare our total runtime with both data-driven RL methods~\cite{lai2022maskplace,shi2024macro,shi2024macro}, WM (WireMask), MP (MaskPlace) and EP (EfficientPlace), and the parameter-tuning method, AutoDMP~\cite{agnesina2023autodmp} on ISPD2005 and MMS benchmarks. 
as shown in \Cref{figs:dse-compare} (a) and (b), under the same case-by-case searching setting. Our method is faster than both approaches. In terms of HPWL performance, we find that RL methods still lag behind analytical approaches, based on our self-implementation with open-source code. Our approach also outperforms AutoDMP with a shorter runtime.

We also provide an algorithm-level DSE method in resourced-constrained scenario.
\Cref{figs:dse-compare} (c) compares our method with SVM, RF (RandomForest)  and XGB (XGBoost) on ISPD2005 benchmark. 
The x-axis shows the HPWL ratio, with BB method as the baseline. Under the same DSE setting with 100 design points selection, our method surpasses all others competitors within the same search time.
\Cref{tab:dse-extract} shows the  improvement of runtime compared with extract method, from which we use only one GPU to simulate resource-constrained scenario.
The runtime of DSE method is largely reduced compared with extract method with less HPWL loss.

\subsection{Observations and Discussions}
\minisection{General Observations}
Due to space constraints, we omit some tables. 
Overall, we find that generating and evolving the initialization algorithm plays the most significant role in improving performance with minimal impact on DREAMPlace runtime. 
\textbf{The key insight is that macro position initialization is critical, often leading significant improvements.} Among the three algorithm components, initializations  have the highest generation success rate, followed by optimizers, with preconditioners performing slightly worse.
%

\begin{figure}[tb!]
    \centering
    \includegraphics[width=1\linewidth]{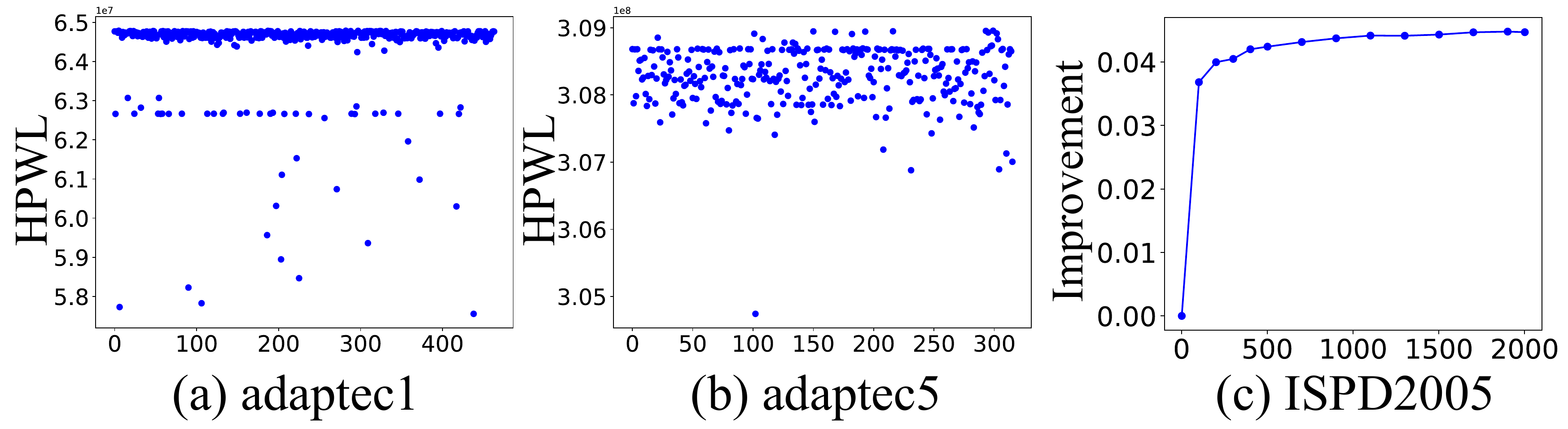}
    \caption{
        Discovered algorithm results observations.
    }
    \label{figs:hpwl-compare}
\end{figure}

\minisection{Discovered Algorithm Observations}
\Cref{figs:hpwl-compare} (a) and (b) illustrate two representative styles of discovered algorithms using the adaptec1 and adaptec5 cases from the MMS benchmark. The x-axis represents the algorithm index. In~\Cref{figs:hpwl-compare} (a), we observe a "pyramid-like" pattern, where groups of algorithms with similar performance cluster together, and fewer algorithms achieve better HPWL as performance improves. In contrast,~\Cref{figs:hpwl-compare} (b) shows an "emergence" phenomenon, where a single standout algorithm significantly outperforms the rest, appearing almost randomly.

\minisection{Inference Scaling Phenomenon}
We also observe an interesting scaling phenomenon regarding the inference scaling law.~\Cref{figs:hpwl-compare} (c) shows the relationship between performance improvement and the number of generated algorithms on the ISPD2005 benchmark. We randomly select different sets of generated algorithms multiple times and average the improvements. Contrary to the conventional expectation of a linear relationship, the curve shows a clear "logarithmic" trend. 
We will provide explorations in future work.

\section{Conclusion}
In this work, we introduce a novel algorithm evolution framework for global placement utilizing large language models.
The framework consists of two main components: an offline generation phase that creates diverse candidate algorithms, and an evolutionary process that combines selection strategies and genetic framework to improve chosen algorithms. 
We also provide an algorithm-level DSE solution targeting resource-constrained scenarios.
The discovered algorithms achieve significant improvements on several benchmarks and demonstrate good generalization ability as well.
We hope this work can provide new insights for this area.

{
\bibliographystyle{IEEEtran}
\bibliography{./references}
}

\end{document}